# Adaptive Resolution Inference (ARI): Energy Efficient Machine Learning for the Internet of Things

Ziheng Wang, Pedro Reviriego, Farzad Niknia, Javier Conde, Shanshan Liu and Fabrizio Lombardi

*Abstract*—The implementation of Machine Learning (ML) in Internet of Things (IoT) devices poses significant operational challenges due to limited energy and computation resources. In recent years, significant efforts have been made to implement simplified ML models that can achieve reasonable performance while reducing computation and energy, for example by pruning weights in neural networks, or using reduced precision for the parameters and arithmetic operations. However, this type of approach is limited by the performance of the ML implementation, i.e., by the loss for example in accuracy due to the model simplification.

In this paper, we present Adaptive Resolution Inference (ARI), a novel approach that enables to evaluate new tradeoffs between energy dissipation and model performance in ML implementations. The main principle of the proposed approach is to run inferences with reduced precision (quantization) and use the margin over the decision threshold to determine if either the result is reliable, or the inference must run with the full model. The rationale is that quantization only introduces small deviations in the inference scores, such that if the scores have a sufficient margin over the decision threshold, it is very unlikely that the full model would have a different result. Therefore, we can run the quantized model first, and only when the scores do not have a sufficient margin, the full model is run. This enables most inferences to run with the reduced precision model and only a small fraction requires the full model, so significantly reducing computation and energy while not affecting model performance. The proposed ARI approach is presented, analyzed in detail, and evaluated using different datasets both for floating-point and stochastic computing implementations. The results show that ARI can significantly reduce the energy for inference in different configurations with savings between 40% and 85%.

## I. INTRODUCTION

Artificial intelligence and energy dissipation are key priorities for future wireless systems, such as 6G [1]. The improvement in energy efficiency of a network is usually

 

achieved by considering also other features such as quality of service; this area is of particular interest also in the context of the Internet of Things (IoT) [2]. The rise of the IoT has enabled the connection of the virtual and physical worlds, in fact, by 2030, it is estimated that there will be nearly 30 billion connected devices [3]. A typical IoT use case involves training Machine Learning (ML) models with data collected by IoT devices, deploying these models, and using the resulting prediction to modify the physical world through actuation devices. As an initial approach, ML models run on central servers, requiring IoT devices to establish a connection to the cloud, raising concerns about security, scalability, or energy efficiency. Hence, the emerging trend is to shift the execution of ML models from the cloud to the edge of the network and to end devices with lower hardware capabilities, giving rise to new paradigms such as edge AI [4] or tinyML [5]. One of the challenges for the future Internet of Things (IoT) is how to implement ML in IoT devices [6]. In contrast to the cloud where the scale of datacenters enables the use of larger and more complex ML models (that can in many cases be run in parallel on several computing units), in the IoT, computing and energy are constrained. This occurs because in many cases, devices have limited hardware and rely on batteries for the energy supply. Hence, ML models must be optimized or compressed when used in IoT [7], [8], [9]; this can be done in many ways, for example, by pruning parameters, by using distillation to create a simpler but functionally similar model, by reducing the bit-width of the parameters and arithmetic operations, or by using alternative energy-efficient implementations such as stochastic computing [8], [10].

In a conventional computing system, each number is represented by a group of bits, for example, 16 or 32. The process of mapping the numbers to a bit representation is known as quantization. The use of more bits provides a more accurate representation of the numbers and the arithmetic operations, but it also requires more memory and power dissipation. Memory is linear with the number of bits, while arithmetic operations (such as multiplication]) are typically polynomial on the number of bits [11]. Therefore, a reduction in the number of bits in an ML model is very attractive for IoT devices [8]. Most general-purpose ML libraries and implementations use floating-point representations with single (32-bit) or half (16-bit) precision that can be reduced with limited loss in performance (e.g., classification accuracy) [12], [13]. The use of shorter floating-point formats (for example of only 8 bits) has been advocated by industry [14], [15].

The implementation of compact neural networks with stochastic computing is also attractive to reduce circuit area and energy [16]. In a stochastic computing system, numbers are represented by sequences of bits that are processed one bit at a time, so sequentially [17]. In this case, the precision typically depends on the length of the sequence, and with longer sequences the computation achieves better results at the cost of more time and energy. The length of the sequences is commonly a power of two, so that each time the length is increased, the processing time and the energy are approximately doubled [10]. Therefore, it is beneficial to use shorter sequences when possible.

The design of quantized ML models is driven by a trade-off between hardware complexity, energy dissipation and model performance [18]. In most cases, the approach followed is to reduce complexity with limited degradation in the model performance; so, this leads to complex algorithms that may require, for example, quantization-aware training [19]. Even with complex algorithms, there is still a small impact on the ML model performance and a limitation on the hardware and dissipation savings as further reductions in hardware would impact the ML model performance.

In this paper, we present Adaptive Resolution Inference (ARI) a novel approach to reduce the energy dissipation of ML inference. ARI is based on the observation that quantization typically introduces only small deviations in the classifier scores and thus, inferences for which the scores have a large margin over the decision boundary are very unlikely to change due to quantization. Therefore, we can run inference on a strongly quantized ML model and check the margin of the scores: if there is sufficient margin, the result is reliable, otherwise, the full model must be run. When the percentage of times that there is a sufficient margin is large, this approach can significantly reduce energy dissipation. ARI can thus potentially achieve the same classification performance as the full model at a fraction of the energy requirement.

The main contributions of the paper are:

1) To propose the combined use of several ML models with different resolutions to reduce energy dissipation compared to a single model.
2) To present ARI, a scheme to combine ML models enabling new energy dissipation versus accuracy tradeoffs in the design of ML systems targeting IoT devices.
3) To evaluate ARI with different datasets and ML model implementations showing its benefits.

The rest of the paper is organized as follows. Section II briefly reviews related work on compressing ML models, presents the problem statement, and describes the ML models that are used in the rest of the paper; section III presents the proposed ARI scheme covering both conventional and stochastic computing implementations. The evaluation results for different datasets and classifiers are presented in section IV that also illustrates the use of ARI in a case study. The paper ends with a conclusion section and a brief discussion of future work.

## II. Preliminaries

This section briefly discusses different optimization techniques proposed to reduce the complexity and energy dissipation of ML models for IoT applications; it also formulates the problem statement by combining different ML models, and finally presents the ML models that are used in the rest of the paper.

### A. Machine Learning Optimizations

The implementation of ML classifiers in IoT devices poses challenges in terms of energy dissipation because many IoT devices run on batteries or solar cells, and therefore, they have a limited energy budget. As discussed in the introduction, there are many approaches to reducing the complexity and energy dissipation of ML classifiers. In the following, some of them are briefly discussed focusing on the ones used in the rest of this paper and for optimization trade-offs.

*1) Model optimization:* One of the approaches to reduce the complexity of ML models is to simplify the models themselves. For example, in many neural networks, there are many weights with very small values. An optimization approach is to prune those parts of the model that do not contribute significantly to its performance [20]; this can significantly reduce the number of parameters of a neural network [21] and the computational effort and energy needed for each inference.

Another optimization approach is knowledge distillation, which transfers knowledge from a complex to a simpler model and has been used successfully for many models and applications [22], [23]. These model optimizations are useful to enable the use of ML in IoT devices; however, in most cases, the energy per inference still limits the number of inferences that can be made.

*2) Resolution optimization:* Another approach to saving the energy dissipation of ML implementations is to reduce the resolution of the model parameters and operations. In the case of conventional computing implementations, the resolution is linked to the format used to represent the numbers. For example, in floating-point representations, half, single, and double precision formats are defined by the IEEE 754 standard with 16, 32, and 64 bits respectively [24]. Instead, for stochastic computing implementations, the resolution is related to the length of the stochastic sequences used to represent numbers [10]. The lengths are typically powers of two because they are generated by maximum-length linear feedback shift registers.

The energy dissipation of a given ML model varies significantly depending on the resolution when using floating-point representations [14]. Similarly, for stochastic computing, as the sequence length increases so does the power dissipation that is approximately linear in the sequence length. Lower resolutions not only significantly reduce the energy dissipation per inference, but they also reduce the circuit delay in floating-point implementations or the number of bits to process in stochastic computing implementations, enabling faster inferences.

*3) Optimization trade-offs:* ML model optimization techniques typically lead to a degradation of performance for example, in terms of classification accuracy. Therefore, in most cases, the strategy followed is to optimize the model subject to an acceptable performance loss. This puts a limit on the optimizations that can be performed, because when reducing the resolution, it must result in a negligible or small performance loss of the ML model.

*B. Problem statement*

In the previous subsection different ML model optimization techniques have been discussed; all of them reduce the complexity of the original model by lowering the resolution, pruning parameters or using distillation to achieve energy saving. In this situation, rather than trying to design new optimizations for ML implementation to meet the requirement of IoT devices for both performance and energy dissipation, we consider an alternative problem: *given a set M of ML classifier models with different energy and performance tradeoffs ranging from the most complex and accurate to the simpler and less accurate: can we combine them to implement a classifier that achieves a target classification performance and has a lower energy dissipation than the models in M that achieve that performance?* The problem is better explained with an example, Consider that we have three models that can meet either a satisfactory performance or energy dissipation as shown in Fig. 1; can we use the three models to build a system that achieves practically the highest performance among the models (as $M_3$) yet lower energy dissipation (as $M_1$ or $M_2$)? This is the goal of ARI: to combine ML models that have been optimized possibly with different techniques to achieve a further reduction in energy dissipation and enable new trade-offs between ML classification performance and energy dissipation.

Therefore, the problem considered is not to further optimize a single ML model but how to combine several models to achieve further energy savings; so, the problem addressed is conceptually different from the classical ML optimization design and builds on top of existing techniques. Hence, it does not make sense to compare the proposed solution with existing optimization techniques as precisely these existing optimized ML models are the inputs to the problem that we are addressing.

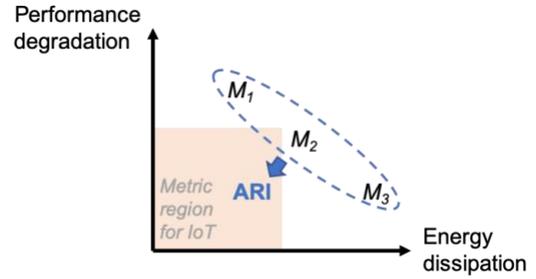

Fig. 1. Problem statement: given a set of ML models with different performance and energy dissipation, can we combine them to design classifiers that provide better metrics for IoT devices?

*C. Machine learning models for evaluation*

To evaluate the proposed scheme, a multilayer perceptron (MLP) has been selected as ML model. This choice has been made to strike a balance between models that could be too simple and others that are too complex to be implemented in IoT devices. Due to its reasonable complexity, the MLP is widely used in IoT applications [25], [26], and at the same time, its structure and operations are like those of other neural networks. Both floating-point and stochastic computing implementations are considered with three widely used datasets: SVHN [27], Fashion MNIST [28], and CIFAR10 [29]. The considered MLP has five layers with neurons of input size-1024-512-256-256-10. The input size changes with the dataset, i.e., 784 for Fashion-MNIST and 3072 for CIFAR10 and SVHN. All designs are implemented in Verilog and mapped to a 32nm library using Cadence Genus for synthesis.

For the floating-point implementations, the full model corresponds to a half-precision (16-bit) floating-point implementation, and different reduced precision models are derived from it by removing the least significant bits from the mantissa as illustrated in Fig. 2. For stochastic computing implementations, the full model uses a sequence length of 4096, and shorter sequences are used for the reduced precision models. The implementation scheme for each type of data representation is as follows.

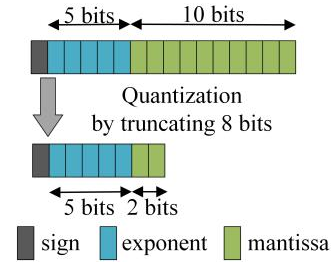

Fig. 2. Example of the reduction of bits in the floating-point evaluation.

TABLE I
AREA AND ENERGY FOR THE FLOATING-POINT MLP WITH DIFFERENT PRECISION FOR FASHION MNIST

| Precision* | Area ($mm^2$) | Energy ($\mu J$) |
|---|---|---|
| FP16 | 0.41 | 0.70 |
| FP14 | 0.34 | 0.57 |
| FP12 | 0.28 | 0.46 |
| FP10 | 0.21 | 0.36 |
| FP8 | 0.14 | 0.25 |

* The precision is denoted as FP followed by the bit width

TABLE II
TIME AND ENERGY PER INFERENCE FOR THE SC MLP WITH DIFFERENT SEQUENCE LENGTHS FOR FASHION MNIST

| Sequence length | Latency ($\mu s$) | Energy ($\mu J$) |
|---|---|---|
| 4096 | 4.10 | 2.15 |
| 2048 | 2.05 | 1.08 |
| 1024 | 1.03 | 0.54 |
| 512 | 0.52 | 0.27 |
| 256 | 0.26 | 0.14 |
| 128 | 0.13 | 0.07 |

*1) Floating-point implementation:* In this case, a hybrid computational scheme that combines serialization and parallelization has been implemented for inference of an MLP; this results in a flexible implementation that is nearly independent of neural network size [30]. In such a scheme, 64 processing elements (PE) are used to calculate multiply-accumulate operations among neurons; each PE includes a MultiplyAccumulate (MAC) unit plus a comparator to implement the ReLU activation function. Then, the MLPs run on an interconnected bank of PE units, and the parameters are stored in SRAMs as shown in Fig. 3. When using the same data precision, the hybrid design requires the same area and power per cycle for all types of datasets and topologies; however, the total latency (number of required cycles multiplied by the critical path delay) and its related energy to complete an inference depend on the network topology of the MLP. For different data precisions, the overheads vary; for example, Table I shows the area overhead and energy dissipation of the design with different bit widths for Fashion MNIST. The cycle duration is 2.51 ns which is the same for all precisions, so the latency in each case is similar. From Table I it can be observed that both area and energy reduce significantly as bits are removed from the mantissa. As per this table, the relative savings on energy per inference can be computed because the time needed for inference does not depend on the precision of the model. For example, reducing from 16 to 10 bits reduces the energy by approximately half. More details on this design are provided in [30].

*2) Stochastic computing implementation:* For stochastic computing, the MLP is implemented in a fully parallel mode such that all neuron values are calculated simultaneously. This is possible because the operations are much simpler, for example, an AND (XNOR) gate can perform the multiplication on unipolar (bipolar) stochastic computing bitstreams, and the result is based on the fraction of 1's/0's on a binary sequence. A bipolar representation with 10-bit precision (1024-bit stochastic sequence) is used in the design. A 10-bit Linear Feedback Shift Register (LFSR) is used for generating a stochastic sequence in the Stochastic Number Generator (SNG) that is used to generate the inputs. The activation function is implemented using a Linear Finite State Machine (LFSM). The diagram of the stochastic computing implementation of a neuron is shown in Fig. 4.

The same circuit can be utilized to run different sequence lengths, so the area overhead is kept the same. However as introduced previously, when using longer sequences, the processing time and the energy per inference of the MLP increase. Table II reports the results for a stochastic computingbased MLP with four layers of 784-100-200-10 for Fashion MNIST [31]; it can be observed that the energy per inference scales almost linearly with the sequence length. Therefore, for stochastic computing, the relative energy savings can be estimated directly from the sequence lengths. The interested reader will find more detailed information about this stochastic computing implementation in [31].

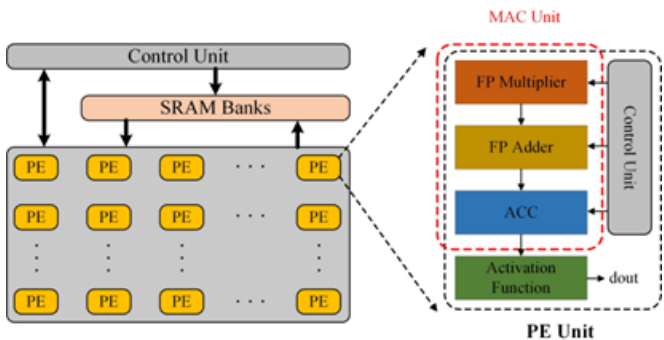

Fig. 3. Block diagram of the floating-point implementation of the MLP.

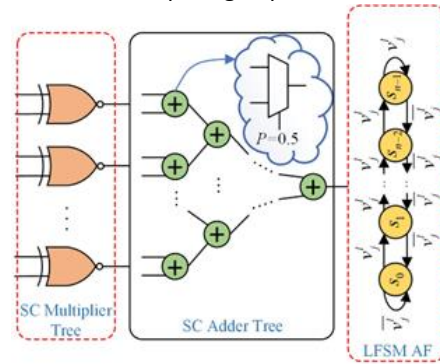

Fig. 4. Diagram of the stochastic computing implementation of a neuron.

## III. Adaptive Resolution Inference (ARI)

This section presents the proposed ARI scheme; first, the motivation and overall approach are described, then the selection of the parameters and energy savings are discussed. Finally, the implementations for both conventional and stochastic computing are analyzed.

### A. Motivation

To understand the rationale behind our proposed scheme, consider an ML classifier, for example, a simple MLP implemented with stochastic computing for the SVHN dataset. As discussed before, in stochastic computing numbers are represented using a sequence of bits that are processed serially, so the larger the length of the sequence the better the resolution. This implies that a higher performance of the MLP can be achieved by utilizing longer sequences, but a larger energy is dissipated.

To better understand the benefit of using longer sequences and the corresponding overhead, Fig. 5 (upper) shows both the classification accuracy and the relative energy per inference (starting with 100% for a sequence length of 128) of the MLP with different sequence lengths. As per the figure, the improvement in accuracy is considerably smaller as the length increases, while the energy per inference of the MLP increases linearly. Therefore, small improvements in accuracy need very large increments in energy dissipation. For example, to gain 0.5% accuracy we must increase the sequence length from 512 to 4096, incurring an 8x increase in energy dissipation. A similar analysis applies to floating-point implementations; a large energy is required for a small improvement in accuracy. Therefore, it is of interest to design schemes that can attain accuracy gains without incurring a large hardware overhead, thus making the relation between accuracy and energy dissipation less steep. This is the main goal of the API scheme as proposed in this paper.

### B. Overall approach

Our goal is to achieve the gain in accuracy of using more bits or large sequence lengths but also reducing the impact on energy dissipation. From the discussion in the previous sections, it is known that a reduced precision model in most cases produces the same inference result as a model with more bits or larger sequences. Therefore, the principle of the proposed scheme is to identify when reduced precision may generate a different result. If this tends to occur with a significant probability, the full model can be executed; else, the reduced precision can be utilized instead to save energy.

Since the outputs of a classifier are the scores for each class, this information is exploited in the proposed scheme to determine whether the inference on a reduced precision model is reliable. Specifically, the difference between the maximum and the second-largest scores is computed and checked. This is illustrated in Fig. 6 for an element in a stochastic computingbased MLP with a sequence length of 4096 bits. The difference between the two highest scores for this element is shown to be very large with a sequence length of 4096. Moreover, our simulation results show that for most elements of the dataset, the difference between the two highest scores is large, in many cases close to one.

Next, consider the effect of reducing the bits or the sequence length on the scores. Quantization has traditionally been modeled as noise when there are many quantized operations that are not strongly correlated, for example in signal processing systems [32]. This is like the case of ML classifiers and thus, we would expect their values to deviate slightly from their original values. The same reasoning applies to sequence lengths in stochastic computing; this is needed, as otherwise the classification accuracy drops significantly, and the reduced precision classifier is not useful. For the MLP discussed above, the reduction of the sequence length to 512 changes the scores for 1.3% (350/26032) of the elements in the dataset. An example of such changes in the scores introduced by reducing precision is shown in Fig. 6; although the difference between the two highest scores is reduced from 0.9844 to 0.8431, they are still associated to the same categories and the score difference is still larger than zero, i.e., the classification result keeps the same (category 6). Therefore, as most of the inferences have a difference between the maximum and the second-largest scores of close to one, quantization of using the reduced length of 512 for this dataset is unlikely to change the classification compared to the full model.

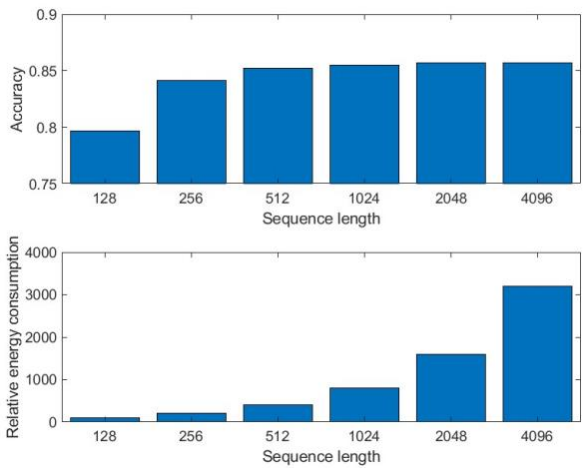

Fig. 5. Accuracy (top) and relative energy per inference (bottom) of a stochastic computing MLP for the SHVN dataset. Accuracy gains are smaller as sequence length increases while energy increases linearly incurring in a large energy use.

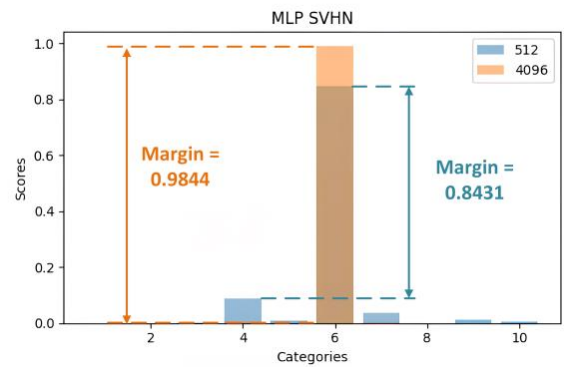

Fig. 6. Example of the classification scores of an element in an MLP for the SHVN dataset with stochastic computing implementation of length 4096 and 512. The classification result (category 6) does not change when reducing the sequence length even though its score is reduced.

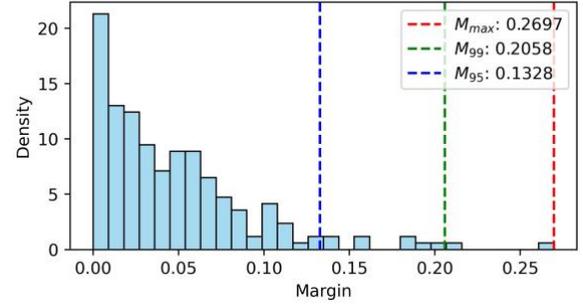

Fig. 8. Distribution of the margin values on the reduced precision model for elements that have different classification results with the full model. Density is computed as the number of elements in the interval divided by the interval size. The values of the threshold for $M_{95}, M_{99}, M_{max}$ are also shown.

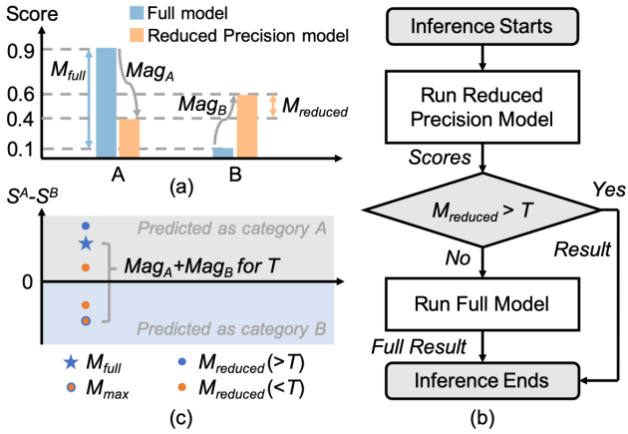

Fig. 7. Proposed Adaptive Resolution Inference (ARI) scheme: (a) an example to show $Mag_A + Mag_B = M_{full} + M_{reduced}$ (0.5 + 0.5 = 0.8 + 0.2) when the reduced precision model changes the classification result from category A to B; (b) the flowchat of ARI; (c) relationship among $M_{reduced}$, $T$, and classification result (when $M_{reduced} > T$, the dot must be above $M_{full}$, which leads to the same category as $M_{full}$; else, it must be a dot between $M_{full}$ and $M_{max}$, which may lead to a different category).

For general purposes, the difference between the maximum and the second-largest inference scores for a given element can be considered as the margin (as marked in Fig. 6) to tolerate errors due to the reduced precision. So, define $M = S^{st} - S^{nd}$ where $S^{st}$ is the largest score among the classes and $S^{nd}$ the second largest. For a given element, when the margin is larger than the worst-case quantization error, the classification result of the quantized model is the same as for the full model. This is conceptually like a scheme proposed to protect K Nearest Neighbor (KNN) classifiers against soft errors [33].

The reliable margin on a reduced precision model (that is large enough to keep the same classification result) can be checked against a threshold $T$. Consider the cases when the reduced precision model changes the classification result; the overall changes on scores caused by the quantization error must have increased the margin on the full model ($M_{full}$) by a magnitude that is exactly the margin on the reduced precision model ($M_{reduced}$). This is illustrated by an example given in Fig. 7 (a); the total magnitude of score changes on the two categories A and B, $Mag_A + Mag_B$, is equal to $M_{full} + M_{reduced}$. However, the impact of quantization error tends to be only minor as discussed previously, because otherwise, the model is not useful. This is also verified by our simulation results; $M_{reduced}$ is observed to be very small for all datasets considered in this paper when the reduced precision model changes the classification result. Therefore, when the margin observed on the reduced precision model is sufficiently large ($> T$), the classification result is still reliable under the quantization errors.

Therefore, as illustrated in Fig. 7 (b), in the proposed scheme the inference is initially run on the reduced precision model. The margin that is computed based on classification scores is then compared against $T$. This threshold is a parameter of the proposed scheme that is discussed in the next subsection. If the margin is above the threshold, the result should be the same as for the full model, so there is no need for further analysis. Instead, when the margin is lower, the result could have been modified by the quantization errors, and thus the inference is run again for this element on the full model. Note that the margin should be computed per element/inference. Overall, if most inferences have a sufficient margin, then the energy of the inference approaches to the one required for the reduced precision model, because the full model is only executed on a small fraction of the inferences.

### C. Parameter selection

The main parameter of the proposed scheme is the threshold $T$; it is used to decide if the difference between the two largest scores is sufficient to assume that the result of the inference in the reduced precision model is the same as for the full model. To determine its value, the inference can be run on the dataset for the full and reduced precision models logging the elements that have a different result in the two models and their margins in the reduced precision model. Then, $T$ can be set to the largest margin value $M_{max}$ among those elements; it also reflects the largest impact of quantization errors because $Mag_A + Mag_B = M_{full} + M_{reduced}$ and $M_{full}$ is fixed. Such a value ensures that the proposed ARI scheme runs the inference again with the full model for all elements that have a different predicted class, and thus generates the same results as the full model for all elements in the dataset. To better explain the setting of $T$, an example for the relationship between $M_{reduced}$ and $T$ is given in Fig. 7 (c). In this example the binary classification between categories A and B is considered; thus, $S^A - S^B$ should be positive (negative) when the predicted category is A (B). The distance between

$M_{full}$ and $M_{max}$ as shown in the figure is the largest value of $Mag_A + Mag_B$, which reflects the largest impact of errors. Therefore, for all possible errors, the position of $M_{reduced}$ cannot be below $M_{max}$ in the figure. This means that when the condition $M_{reduced} > T$ is met in the proposed ARI scheme, the classification result should be the same as that in the full model (i.e., category A), because $M_{reduced}$ can only be within the positive region of $S^A - S^B$ as shown in Fig. 7 (c). Else, when $M_{reduced} > T$ is not met, $M_{reduced}$ can possibly be within the negative region of $S^A - S^B$, predicting the category as B; in this case, ARI runs the full model to ensure that the classification result of category A is obtained.

Assuming that the dataset is representative of the elements found in the field, this selection of $T$ ensures that ARI generates almost the same classification results as the full precision model. If additional reductions in energy per inference are needed, $T$ can be computed as the margin of a given percentile of the elements that changed their classification result when using the reduced precision model. For example, consider the margin that covers 99% of the elements that changed their classification result (denoted as $M_{99}$). This value would ensure that only for 1% of the elements that could have changed the classification result, the full model is not run. In the example discussed previously, there were 350 such elements; thus, only 3 elements could produce a different result leading the worst case to 0.0115% (3/26032) accuracy loss. This may be acceptable depending on the application if it enables further energy savings. As an example, the distribution of the margin over the 350 elements that change the classification results, is illustrated in Fig. 8 showing the thresholds $T$ that correspond to $M_{95}$, $M_{99}$ and $M_{max}$. The margin can be significantly reduced (from 0.2697 to 0.2058 or 0.1328 for $M_{99}$ or $M_{95}$) by leaving only a small percentage of the elements that change their classification result uncovered by the threshold.

In summary, from the dataset used to train the ML classifier, the values of $T$ can be calculated to achieve the desired performance in terms of classification accuracy with the proposed ARI scheme: so using $M_{max}$ to achieve the same accuracy as the full model and using $M_{xy}$ to ensure that for xy% of the elements that change the classification results, the full model is run, leaving only a small fraction undetected. Note, that there is a trade-off when selecting the value for $T$ as larger values imply a larger fraction of inferences that need to run the full model and thus, requiring more energy. This is analyzed in detail in the evaluation section.

### D. Energy per inference

To estimate the energy per inference of the proposed scheme, define the fraction of inferences that must run the full model as $F$ and the energy per inference of the reduced precision and full model as $E_R$ and $E_F$ respectively. Then the average energy per inference when using ARI is approximately:

$$E_{ARI} = E_R + F \cdot E_F. \qquad (1)$$

As discussed previously, we expect $F$ to be significantly smaller than one so that only a fraction of the inferences must be executed on the full model and $E_R$ to be also significantly smaller than $E_F$. For example, consider $F = 0.2, E_R = 0.25$ and $E_F = 1$, then $E_{ARI} = 0.45$, hence, less than half of the energy required by the full model. The potential savings are assessed in the evaluation section using equation (1).

### E. Implementation

As discussed previously, the main goal of the proposed ARI scheme is to achieve the same (or nearly the same) classification accuracy as the full model at a fraction of the energy required. In general, ARI requires the use of two ML models, full and reduced precision; however, the reduced precision model is significantly less complex. The implementation scheme that supports these two models is explained next for floating-point and stochastic computing MLPs respectively.

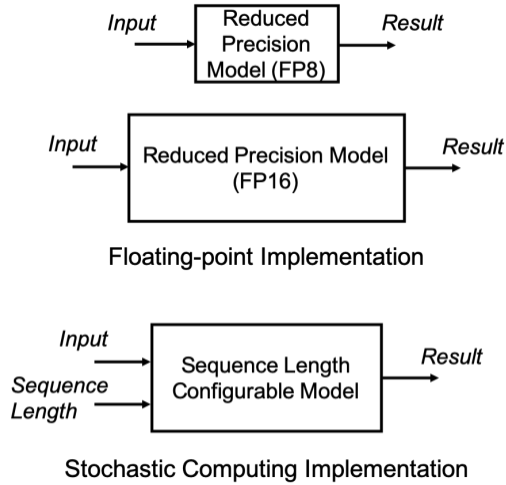

Fig. 9. Proposed implementations: floating-point (upper) and stochastic computing (lower).

*1) Floating-point:* for the floating-point implementation, the use of two implementations is proposed. For example, a full model with a 16-bit floating-point format (FP16) and a reduced precision with an 8-bit format (FP8). This is illustrated in Fig. 9 (upper part). The circuit area of the reduced precision model is only a fraction of the full precision model, and its overhead is likely to be offset by the energy savings. The use of independent models is needed because implementing in hardware a model that supports both resolutions is complex and introduces additional overheads.

*2) Stochastic Computing:* for stochastic computing, the implementation of a model that supports different sequence lengths is easier and, therefore, feasible, because most blocks are independent of the sequence length. Therefore, the implementation considered is a single model that can be configured with different sequence lengths as shown in Fig. 9 (lower part). The overhead of supporting the two models is negligible.

## IV. Evaluation

This section presents the results of the evaluation of ARI on the ML models, implementations, and datasets described in section II-C. Initially, the margin and thresholds when using the proposed ARI are analyzed; then, the fraction of inferences that need to run the full model is evaluated, and the two main performance metrics for the application: energy per inference and accuracy are assessed. Finally, a case study is presented to illustrate the potential benefits of ARI, showing a reduction in the energy consumption with no accuracy loss on the dataset.

The proposed scheme has been evaluated with three widely used datasets: SVHN, Fashion MNIST, and CIFAR10 and a Multilayer Perceptron (MLP) as ML model. The MLP has five layers with sizes input-1024-512-256-256-10. The input size changes with the dataset, 784 for "fashion-MNIST" and 3072 for "CIFAR10" and "SVHN". The Parametric Rectified Linear Unit (PReLU) is selected as the activation function. The model is pre-trained as the full precision model in 20 epochs with format FP16 for the floating-point implementation and with sequence length 4096 for the stochastic computing implementation.

### A. Margin analysis and threshold selection

Consider initially the margin for the different datasets and MLP implementations with different degrees of quantization. The margin of the elements that have different classification results in the reduced precision model, is shown in Fig. 10 for floating-point and in Fig. 11 for the stochastic computing implementations for the three datasets. As more bits (or sequence length) are removed, the values of the margin for the elements that change the classification results, increase.

As per Fig. 10, for floating-point implementations the removal of 4 bits only impacts elements with small margins (at most 0.0609) for all datasets and threshold values. Removing 6 bits increases the margin to larger values, in the 0.1 to 0.3 range depending on the dataset and threshold value. Finally, removing 8 bits leads to larger margin values that are in the 0.6 to 0.8 range for $M_{max}$ and can be reduced to 0.4 to 0.6 by using $M_{99}$ and to 0.3 to 0.4 by using $M_{95}$. In summary, the analysis of the margin suggests that removing 4 bits requires a small fraction of inferences with the full model, and removing 6 bits may also be acceptable; removing 8 bits may require the use of the $M_{95}$ or $M_{99}$ threshold values to achieve a low number of recomputations for the full model.

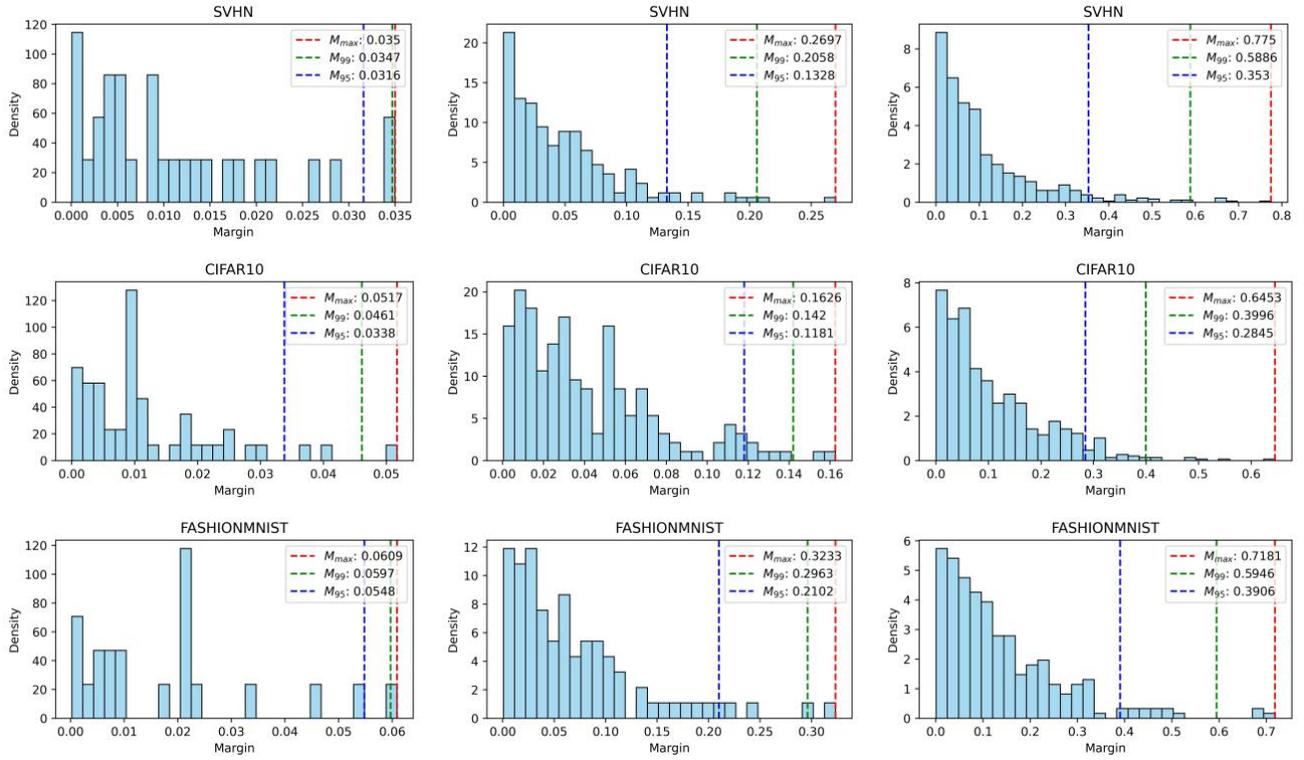

Fig. 10. Distribution of the margin values for elements that have different classification results with the full and reduced precision floating-point MLP models when removing 4 (left), 6 (center), and 8 (right) bits from the mantissa on the SVHN (first row), CIFAR-10 (second row) and Fashion MNIST (third row).

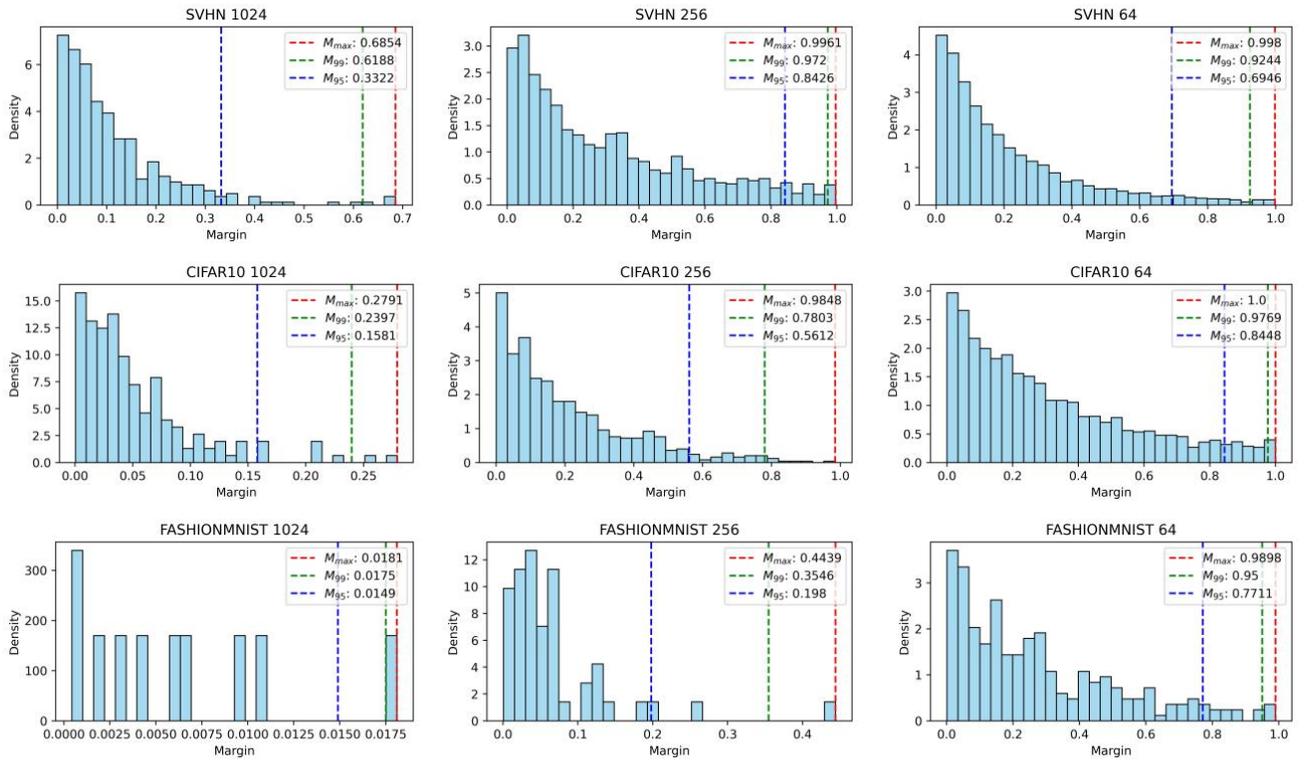

Fig. 11. Distribution of the margin values for elements that have different classification results with the full and reduced precision stochastic computing MLP models with sequence lengths of 1024 (left), 256 (center), and 64 (right) on the SVHN (first row), CIFAR-10 (second row) and Fashion MNIST (third row).

For the stochastic computing implementations, the margin values are highly dependent on the dataset as shown in Fig. 11. Reducing the sequence length affects elements with large margin values for SVHN, somewhat smaller values for CIFAR-10, and smaller values for Fashion MNIST. For SVHN, reducing the sequence length to 1024 (from its original length of 4096) already leads to large values of the thresholds that can be reduced to 0.33 by using $M_{95}$. For CIFAR10, the threshold values are reduced to the 0.15 to 0.3 range; reducing the sequence length to 256 leads to large values, above 0.5 for all cases. Finally, in Fashion MNIST, the sequence length can be reduced even to 256 with values in the range of 0.2 to 0.45. Therefore, it seems the sequence length can be reduced by a factor of 4x when using ARI in all datasets and even to 256 for Fashion MNIST.

After analyzing the margin, the selection of $T$ is next considered. The values for $T = M_{max}$ and those obtained by selecting the margin that corresponds to the 95% and 99% largest margins $T = M_{95}$, $T = M_{99}$ are also shown in Figs. 10 and 11. It can be observed that leaving out a small percentage of the elements that change the classification results (1% to 5%), can significantly reduce the threshold. To better visualize the results, Fig. 12 shows the thresholds for the floatingpoint and the stochastic computing implementations with the different datasets as a function of the number of bits removed or the reduction of the sequence length.

For the floating-point implementation, the plots in Fig. 12 confirm that the thresholds start to be significant only after removing 6 bits. The relative values of $M_{max}, M_{99}, M_{95}$ show that the benefit of using the lower thresholds is significant when 6 or more bits are removed. This suggests that additional reductions in energy can be traded for a small loss in accuracy. In summary, the values of the threshold confirm that it is possible to remove a significant number of bits.

For the stochastic computing implementation, the results are different for each dataset. For SVHN, the values start to be significant even when the length is reduced to 2048; for CIFAR 10 for 1024, and for Fashion MNIST for 256. Therefore, the potential reductions in sequence length will be different for each dataset.

*B. Inferences with full model*

Once the value of $T$ is selected, we can also estimate the fraction of inferences that must be executed with the full model due to insufficient margin. This is illustrated in Fig. 13.

This allows, for floating-point implementations to establish the percentage of inferences that need to use the full model; it is below 20% even when removing 6 bits for all datasets; for SVHN and Fashion MNIST we can even remove 7 bits. When using the lower thresholds, the reduction can go up to 8 bits while keeping the fraction of recomputations below 20%. For stochastic computing implementations, the sequence length can be reduced by at least $2X$ requiring less than 20% computations of the full model for all datasets. For Fashion MNIST, the reduction can be $32X$. This clearly shows the potential benefits of the proposed scheme.

*C. Energy savings*

To find the energy savings obtained when using ARI, the energy per inference values of the implementations described in section II-C are used such that the energy per inference of the full ($E_F$) and reduced model ($E_R$) can be obtained. Then the ratio $\frac{E_R}{E_F}$ is computed and used in conjunction with the fraction of inferences that need to run the full model ($F$) obtained in the simulations to estimate the energy savings as

$$1 - \frac{E_{ARI}}{E_F} = (1-F) - \frac{E_R}{E_F}. \quad (2)$$

The results are shown in Fig. 14. As bits are removed in the reduced precision model, energy savings start to increase, then they reach the maximum, and start to decrease. This is because as we reduce bits, $E_R$ is also reduced, but $F$ increases which at some point reduces the savings.

For floating-point implementations, the potential energy savings are in the range of 40% to 50%. When a threshold value of $M_{max}$ is used, approximately 40% savings are obtained when reducing 6 bits. Using $M_{99}$ or $M_{95}$ enables savings of approximately 45% and 50% for SHVN and Fashion MNIST respectively, instead for CIFAR10 the savings are almost the same as when using $M_{max}$.

For stochastic computing implementations, the potential savings are larger and more dependent on the dataset. The lowest savings are obtained for CIFAR10 and are in the 40% to 60% range depending on the value of the threshold used; in all cases, the sequence length that maximizes the savings is 1024, so a 4x reduction on the 4096 of the full model. For SHVN, savings are in the 55% to 70% range with sequence lengths of 1024 or 512 depending on the threshold value. Finally, for Fashion MNIST, savings are in 80% to 85% range with shorter sequence lengths of 512 or 256. These results confirm that ARI can consistently reduce the energy per inference needed for different implementations and datasets.

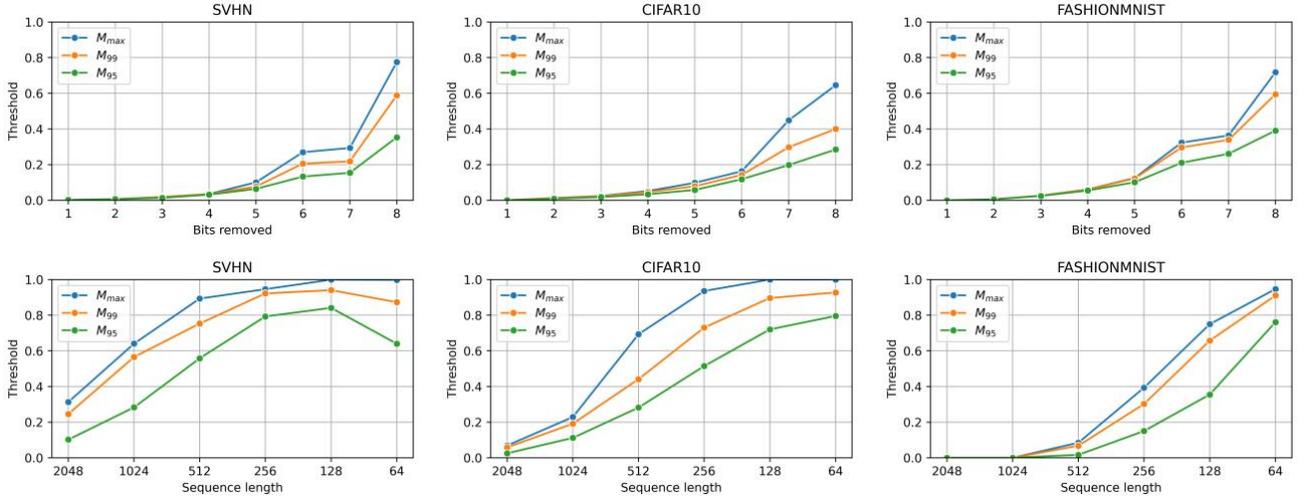

Fig. 12. Margin thresholds for the floating-point (top) and stochastic computing (bottom) with SVHN (left), CIFAR-10 (center), and Fashion MNIST (right).

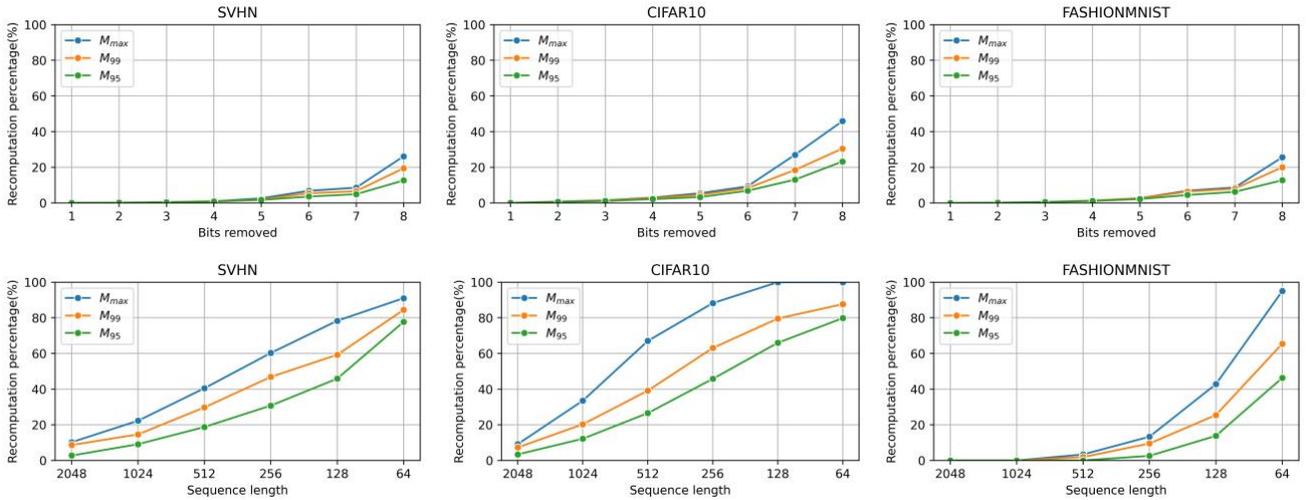

Fig. 13. Fraction of the elements of the dataset that need to run the full model due to insufficient margin for the floating-point (top) and stochastic computing (bottom) with SVHN (left), CIFAR-10 (center), and Fashion MNIST (right).

### D. Accuracy

To conclude the evaluation, it is of interest to assess the impact of ARI on the classification accuracy of the ML model. Obviously, when the threshold $T$ is selected to be the maximum margin $M_{max}$ observed over the entire dataset, the accuracy should be the same as that of the full model. However, when $T = M_{99}$ or $T = M_{95}$, there is a small loss in accuracy. Fig. 15 shows the accuracy for the floating-point and stochastic computing ARI implementations. The losses for the original quantized MLPs (on which the number of bits is reduced directly without employing ARI) are also given for reference; the accuracy drop is minimal and significantly smaller than for the original implementation. Therefore, the use of $T = M_{99}, M_{95}$ can be of interest to obtain additional energy savings with a small impact on accuracy.

From Figs. 14 and 15, a designer can select the most suitable trade-off for the system in terms of accuracy versus energy dissipation. For example, if the acceptable accuracy loss versus the full model is known it can then be used to select the combinations that meet that target from Fig. 15 and then energy savings can be computed from Fig. 14. This is further illustrated in the case study presented in the next subsection.

### E. Case study

To illustrate the use of ARI, we consider as a case study the design of a system that achieves exactly the same classification accuracy in the dataset as the full model (in our case FP16 for floating point and sequence length 4096 for stochastic computing). In the ML model implementations used, reducing precision or sequence length generates an accuracy loss, and thus the traditional quantization schemes cannot be used to reduce energy dissipation without impacting accuracy. Instead, with ARI we can use a threshold value equal to $M_{max}$ with no accuracy loss on the dataset to achieve the energy savings shown in Tables III and IV; significant savings, approximately 40% for the floating point implementations and between 50 and 80% for stochastic computing, are obtained. This illustrates how ARI can combine ML models optimized with existing techniques to achieve further savings.

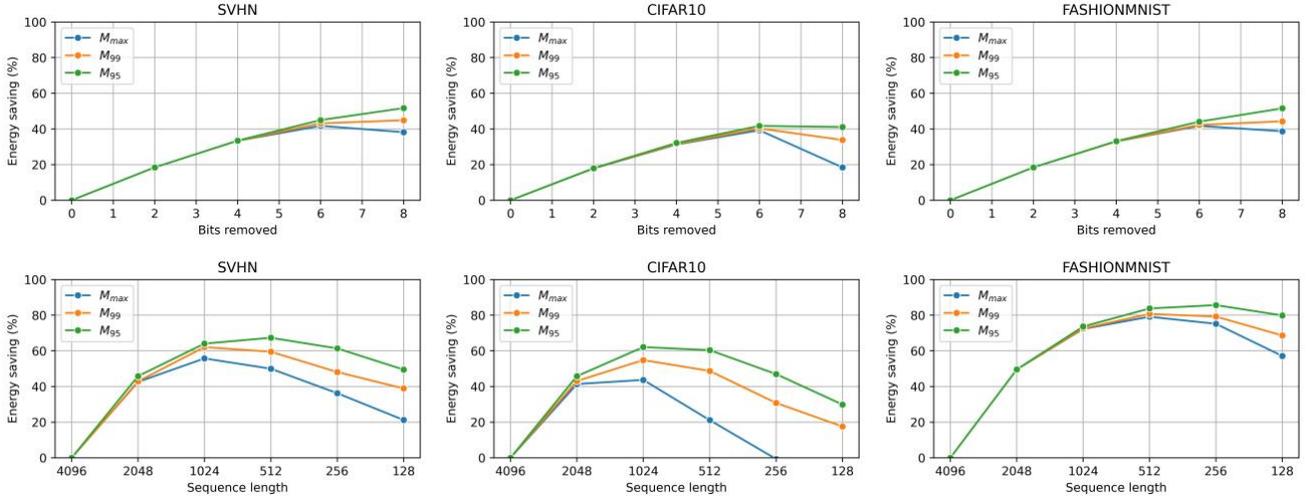

Fig. 14. Energy savings when using the proposed ARI scheme with the floating-point (top) and stochastic computing (bottom) implementations for SVHN (left), CIFAR-10 (center), and Fashion MNIST (right).

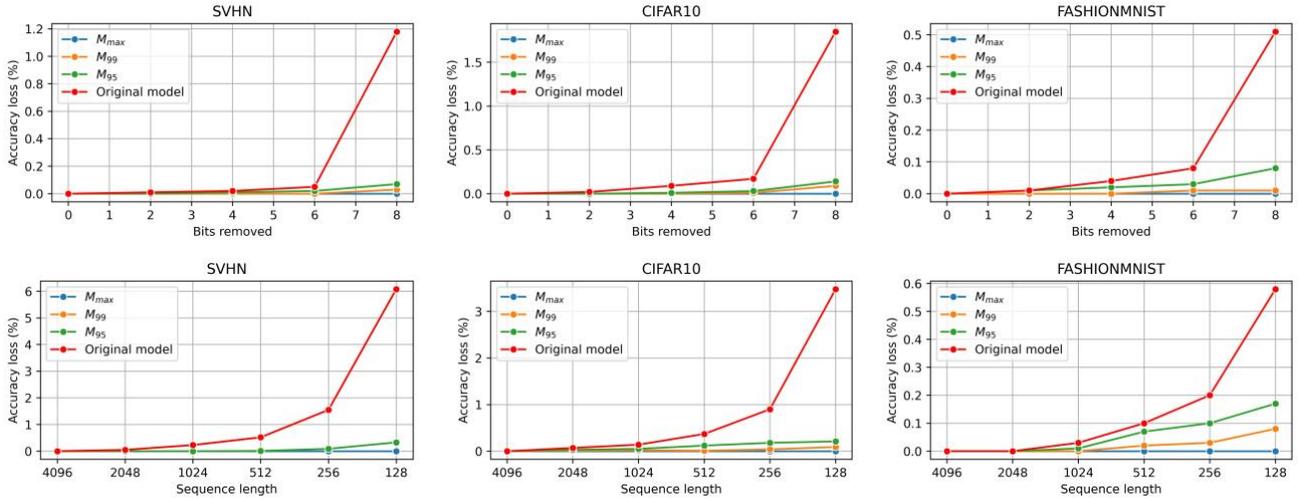

Fig. 15. Accuracy drop in % when using the proposed ARI scheme with the floating-point (top) and stochastic computing (bottom) implementations for SVHN (left), CIFAR-10 (center), and Fashion MNIST (right). The accuracy drop of the original implementation is also shown for reference.

## V. CONCLUSION

In this paper, we have presented Adaptive Resolution Inference (ARI), a scheme to achieve nearly the same classification accuracy of large precision ML models with a fraction of their energy dissipation per inference. This has been achieved by running a reduced precision model and checking if either the result is reliable or it can be modified due to quantization. Only in these cases, the full model is run to generate the original results. ARI has been evaluated on floating-point and stochastic computing implementations of MLPs for several datasets. The simulation results have shown that most inferences use only the reduced precision model, thus significantly reducing energy; specifically, reductions of 40% to 85% in energy per inference are obtained. These results confirm the potential benefits of ARI when implementing ML classifiers in IoT devices.

TABLE III
ENERGY SAVINGS FOR THE FLOATING-POINT MLP WHEN USING ARI WITH NO ACCURACY LOSS ON THE DATASET

| Dataset | Quantization | Savings |
|---|---|---|
| SHVN | FP10 | 41.18% |
| CIFAR10 | FP10 | 39.27% |
| FASHIONMNIST | FP10 | 41.72% |

TABLE IV
ENERGY SAVINGS FOR THE STOCHASTIC COMPUTING MLP WHEN USING ARI WITH NO ACCURACY LOSS ON THE DATASET

| Dataset | Sequence length | Savings |
|---|---|---|
| SHVN | 1024 | 55.76% |
| CIFAR10 | 1024 | 47.70% |
| FASHIONMNIST | 512 | 79.13% |

## VI. Acknowledgements

Pedro Reviriego and Javier Conde would like to acknowledge the support of the FUN4DATE (PID2022-136684OBC22) project funded by the Spanish Agencia Estatal de Investigacion (AEI) 10.13039/501100011033.